\documentclass{article}
\usepackage{spconf,amsmath,bm,cite,graphicx} %
\usepackage{makecell,booktabs,arydshln}

\usepackage{color}
\usepackage{multirow}
\usepackage{etoolbox,siunitx}
\robustify\bfseries
\usepackage{balance}

\makeatletter
\def\adl@drawiv#1#2#3{%
        \hskip.5\tabcolsep
        \xleaders#3{#2.5\@tempdimb #1{1}#2.5\@tempdimb}%
                #2\z@ plus1fil minus1fil\relax
        \hskip.5\tabcolsep}
        
\newcommand{\cdashlinelr}[1]{%
  \noalign{\vskip\aboverulesep
           \global\let\@dashdrawstore\adl@draw
           \global\let\adl@draw\adl@drawiv}
  \cdashline{#1}
  \noalign{\global\let\adl@draw\@dashdrawstore
           \vskip\belowrulesep}}
           
\newcommand{\myhdashline}{%
  \noalign{\vskip\aboverulesep
           \global\let\@dashdrawstore\adl@draw
           \global\let\adl@draw\adl@drawiv}
  \hdashline
  \noalign{\global\let\adl@draw\@dashdrawstore
           \vskip\belowrulesep}}
\makeatother

\title{Semi-Supervised Speech Recognition\\via Graph-based Temporal Classification}
\name{Niko Moritz, Takaaki Hori, Jonathan Le Roux}
\address{Mitsubishi Electric Research Laboratories (MERL), Cambridge, MA, USA} %

\begin{document}
\ninept

\maketitle

\begin{abstract}

Semi-supervised learning has demonstrated promising results in automatic speech recognition (ASR) by self-training using a seed ASR model with pseudo-labels generated for unlabeled data. The effectiveness of this approach largely relies on the pseudo-label accuracy, for which typically only the 1-best ASR hypothesis is used. However, alternative ASR hypotheses of an $N$-best list can provide more accurate labels for an unlabeled speech utterance and also reflect uncertainties of the seed ASR model. In this paper, we propose a generalized form of the connectionist temporal classification (CTC) objective that accepts a graph representation of the training labels. The newly proposed graph-based temporal classification (GTC) objective is applied for self-training with WFST-based supervision, which is generated from an $N$-best list of pseudo-labels. In this setup, GTC is used to learn not only a temporal alignment, similarly to CTC, but also a label alignment to obtain the optimal pseudo-label sequence from the weighted graph. Results show that this approach can effectively exploit an $N$-best list of pseudo-labels with associated scores, considerably outperforming standard pseudo-labeling, with ASR results approaching an oracle experiment in which the best hypotheses of the $N$-best lists are selected manually.

\end{abstract}
\vspace{-0.1cm}
\section{Introduction}
\label{sec:intro}
\vspace{-0.1cm}

Over the last decade, automatic speech recognition (ASR) technologies have progressed to a point where ASR has become a central user interface for various electronic devices.
This progress can largely be attributed to improved acoustic models using more sophisticated neural network architectures, as well as to improved sequence-to-sequence modeling approaches, known as end-to-end ASR, which have greatly contributed to optimizing the training and decoding process of ASR systems \cite{Li2020fast,MoritzHLR20,wang2020low}.
However, to achieve state-of-the-art performance, end-to-end ASR models are generally more data hungry compared to traditional hybrid HMM-DNN ASR solutions~\cite{Pundak2016LFR_NN}.
Although the available amount of manually transcribed training data has grown over the years, a lack of training data still exists, especially for low-resource domains and languages.
To alleviate these problems, data augmentation as well as self- and semi-supervised learning can be applied in order to utilize unlabeled data for training.
In self-supervised learning, typically a latent representation of speech is learned from unlabeled data to pre-train a neural network model for a following supervised or semi-supervised learning step \cite{Baevski2020wav2vec,Liu2020mockingjay,khurana2020cstnet}.
Semi-supervised learning approaches exploit some amount of labeled data to support learning from unlabeled data \cite{Hori2019cyclecon,liu2019adversarial}.
Self-training uses a seed ASR model, trained from transcribed data, to generate pseudo-labels for unlabeled data for further training \cite{Lamel2002lightlySupASR,Novotney2009unsupAMLM,Huang2013SemisupervisedGA,vesely2013semisup}.
This approach has recently become increasingly popular in end-to-end ASR as it has shown promising results \cite{Li2019SemiSupWeakDistil,weninger2020semisupervised,xu2020iterative,khurana2020unsupervised}.
However, as classical pseudo-labeling approaches only exploit the best ASR hypothesis for an unlabeled utterance, they ignore pseudo-label uncertainties of the system that may be useful \cite{vesely2013semisup} as well as alternative hypothesis with potentially fewer errors.
In \cite{hsu2020lpriormatch}, an $N$-best list of ASR hypotheses is used by summing over the weighted losses of multiple pseudo-labels for a single speech utterance, where weights are estimated from scores of a strong language model (LM). In \cite{dey2019exploiting}, multiple pseudo-labels are generated for each unlabeled speech utterance using different dropout settings, which are used for self-training with the purpose of capturing ASR uncertainties.

In this work, we propose a new objective function termed \textit{graph-based temporal classification (GTC)}, which generalizes the popular connectionist temporal classification (CTC) loss function \cite{GravesFGS06} to accept a graph representation as an input for modeling user-defined output symbol structures, including the possibility to assign probabilities to the generated output symbol sequences through transition weights. The proposed GTC loss function can be used to model all possible variations of CTC, including Gram-CTC \cite{liu2017gramctc}, the automatic segmentation criterion (ASG) \cite{collobert2016wav2letter}, and other still unexplored structures that can be modeled using weighted finite automata.

We apply the GTC loss to the semi-supervised learning problem, where we generate a WFST-based graph from an $N$-best list of pseudo-labels in order to leverage the information contained in alternative hypotheses as well as in their ASR scores. In such a setup, GTC is expected to find not only the best temporal alignments, similar to CTC, but also an optimal label sequence encoded in such a graph.
Note that self-training with lattice-based supervision was also proposed in \cite{manohar2018semi,Tong2019unbiasedSemiSup,Sheikh2020} using a hybrid ASR system and the LF-MMI objective in order to incorporate frame-level confidence scores and alternate pseudo-labels. %
However, we here consider the more general case of alternative alignment-free transcriptions using an end-to-end ASR system and the GTC objective function. 

ASR results %
show that the proposed GTC-based semi-supervised learning approach can effectively utilize an $N$-best list of ASR hypotheses for self-training by achieving considerable improvements over the commonly used 1-best pseudo-labeling approach.

\vspace{-0.1cm}
\section{graph-based temporal classification}
\label{sec:gct}
\vspace{-0.1cm}

Let us consider a feature sequence $X$ of length $T'$ derived from a speech utterance, processed by a neural network to output a posterior probability distribution sequence $Y=(\bm y^1,\dots,\bm y^T)$, of length $T$ potentially different from $T'$ due to downsampling, where $\bm y^t$ denotes the vector of posterior probabilities and $y^t_k$ the posterior probability for output symbol $k$ at time $t$. We assume that the labeling information for training is represented by a graph $\mathcal{G}$.
The graph-based temporal classification (GTC) objective function marginalizes over all possible node sequences that can be output by graph $\mathcal{G}$, which includes all valid node patterns as well as all valid temporal alignment paths, the latter being known from CTC \cite{GravesFGS06}. Thus, the conditional probability for a given graph $\mathcal{G}$ is defined by the sum over all node sequences in $\mathcal{G}$, which can be written as:
\begin{equation}
\label{eq:gtc}
    p(\mathcal{G}|X) = \sum_{\pi \in \mathcal{S}(\mathcal{G},T)} p(\pi | X),
\end{equation}
where $\mathcal{S}$ represents a search function that unfolds $\mathcal{G}$ to all possible node sequences of length $T$ (not counting non-emitting start and end nodes), $\pi$ denotes a single node sequence and alignment path, and $p(\pi|X)$ is the posterior probability for the path $\pi$ given feature sequence $X$. 

We introduce a few more notations that will be useful to derive $p(\mathcal{G}|X)$. 
We index the nodes using $g=0,\dots,G+1$, sorting them in a breadth-first search manner from $0$ (non-emitting start node) to $G+1$ (non-emitting end node). 
We denote by $l(g)$ the output symbol observed at node $g$,
and by $W_{(g,g')}$ the transition weight on edge ($g$, $g'$).
Finally, we denote by $\pi_{t:t'}=(\pi_t,\dots,\pi_{t'})$ the node sub-sequence of $\pi$ from time index $t$ to $t'$.
Note that $\pi_0$ and $\pi_{T+1}$ correspond to the non-emitting start and end nodes $0$ and $G+1$.

In CTC, the conditional probabilities $p(\bm l | X)$ for a given label sequence $\bm l$ are computed efficiently by a dynamic programming algorithm, which is based on computing the forward and backward variables and stitching both together to compute $p(\bm l | X)$ at any given time index $t$ \cite{GravesFGS06}.
In a similar fashion, the GTC forward probability can be computed for $g=1,\dots,G$ using
\begin{equation}
\label{eq:grapctc_fw_probs}
    \alpha_t (g) = \sum_{\substack{\pi \in \mathcal{S}(\mathcal{G},T):\\ \pi_{0:t} \in \mathcal{S}(\mathcal{G}_{0:g},t)}} \prod_{\tau=1}^t 
    W_{(\pi_{\tau-1},\pi_\tau)}    y_{l(\pi_{\tau})}^{\tau},
\end{equation}
where $\mathcal{G}_{0:g}$ denotes the sub-graph of $\mathcal{G}$ starting at node $0$ and terminating at node $g$. 
The sum is taken over all possible $\pi$ whose sub-sequence up to time index $t$ can be generated in $t$ steps from the sub-graph $\mathcal{G}_{0:g}$.
The backward variable $\beta$ is computed similarly for $g=1,\dots,G$ 
using
\begin{equation}
\label{eq:grapctc_bw_probs}
    \beta_t (g) = \sum_{\substack{\pi \in \mathcal{S}(\mathcal{G},T):\\ \pi_{t:T+1} \in \mathcal{S}(\mathcal{G}_{g:G+1},T-t+1)}} \prod_{\tau=t}^T 
    W_{(\pi_{\tau},\pi_{\tau+1})}    y_{l(\pi_{\tau})}^{\tau},
\end{equation}
where $\mathcal{G}_{g:G+1}$ denotes the sub-graph of $\mathcal{G}$ starting at node $g$ and terminating at node $G+1$. 
By using the forward and backward variables, the probability function $p(\mathcal{G}|X)$ can be computed for any $t$ by summing over all $g$:
\begin{equation}
\label{eq:fw_bw}
    p(\mathcal{G}|X) = \sum_{g \in \mathcal{G}} \frac{\alpha_t(g) \beta_t(g)}{y^{t}_{l(g)}} .
\end{equation}

For gradient descent training, the loss function
\begin{equation}
\label{eq:loss}
    \mathcal{L} = -\ln{p(\mathcal{G}|X)}
\end{equation}
must be differentiated with respect to the network outputs, which can be written as:
\begin{equation}
\label{eq:deriv_ln_0}
    -\frac{\partial \ln{ p(\mathcal{G}|X)} }{\partial y_{k}^{t}}  = -\frac{1}{p(\mathcal{G}|X)} \frac{\partial p(\mathcal{G}|X)}{\partial y_{k}^{t}},
\end{equation}
for any symbol $k \in \mathcal{U}$, where $\mathcal{U}$ denotes a set of all possible output symbols.

Because ${\alpha_t(g) \beta_t(g)} / {y_{l(g)}^{t}} $ is proportional to $y_{l(g)}^{t}$, 
\begin{equation}
\label{eq:deriv_ln}
    \frac{\partial ({\alpha_t(g) \beta_t(g)}/ {y_{l(g)}^{t}})}{\partial y_{l(g)}^{t}} = \frac{\alpha_t(g) \beta_t(g)} {{y_{l(g)}^{t}}^2},
\end{equation}
and from \eqref{eq:fw_bw}, we can derive
\begin{equation}
\label{eq:deriv_pGX_y}
    \frac{\partial p(\mathcal{G}|X)}{\partial y_{k}^{t}}  = \frac{1}{{y_{k}^{t}}^2} \sum_{g \in \operatorname{\Psi}(\mathcal{G},k)} \alpha_t(g) \beta_t(g) ,
\end{equation}
where $\operatorname{\Psi}(\mathcal{G},k) = \{g \in \mathcal{G}: l(g) = k\}$ denotes the set of nodes in $\mathcal{G}$ at which symbol $k$ is observed.

To backpropagate the gradients through the softmax function, we need the derivative with respect to the unnormalized network outputs $u_k^t$ before the softmax is applied, which is
\begin{equation}
\label{eq:deriv_pGX_u}
    -\frac{\partial \ln{p(\mathcal{G}|X)}}{\partial u_k^t} = - \sum_{k' \in \mathcal{U}} \frac{\partial \ln{p(\mathcal{G}|X)}}{\partial y_{k'}^{t}} \frac{\partial y_{k'}^t}{\partial u_k^t}.
\end{equation}
By substituting (\ref{eq:deriv_pGX_y}) and the derivative of the softmax function $ \partial y_{k'}^t / \partial u_k^t = y_{k'}^t \delta_{k k'}  -  y_{k'}^t y_k^t $ into (\ref{eq:deriv_pGX_u}), we finally derive 
\begin{equation}
    -\frac{\partial \ln{ p(\mathcal{G}|X)} }{\partial u_k^t}  = y_k^t - \frac{1}{y_k^t p(\mathcal{G}|X)} \sum_{g \in \operatorname{\Psi}(\mathcal{G},k)} \alpha_t(g) \beta_t(g) ,
\end{equation}
where we used the fact that
\begin{align}
    \sum_{k' \in \mathcal{U}} \frac{1}{{y_{k'}^{t}}} \sum_{g \in \operatorname{\Psi}(\mathcal{G},k')} \alpha_t(g) \beta_t(g) &= \sum_{k' \in \mathcal{U}} \sum_{g \in \operatorname{\Psi}(\mathcal{G},k')}  \frac{\alpha_t(g) \beta_t(g)}{{y_{l(g)}^{t}}}  \nonumber \\
    &= \sum_{g \in \mathcal{G}}  \frac{\alpha_t(g) \beta_t(g)}{y^t_{l(g)}} \nonumber \\
    &= p(\mathcal{G}|X),
\end{align}
and %
\begin{align}
    \sum_{k' \in \mathcal{U}} \frac{\partial \ln{p(\mathcal{G}|X)}}{\partial y_{k'}^{t}} y_{k'}^t y_k^t =  \frac{1}{p(\mathcal{G}|X)} {p(\mathcal{G}|X)}  y_k^t = y_k^t.
\end{align}

For efficiency reason, we implemented the GTC objective in CUDA as an extension for PyTorch.

\vspace{-0.1cm}
\section{Graph Generation for Self-Training}
\label{sec:graph_generation}
\vspace{-0.1cm}

\begin{figure}[tb]
  \includegraphics[width=\linewidth]{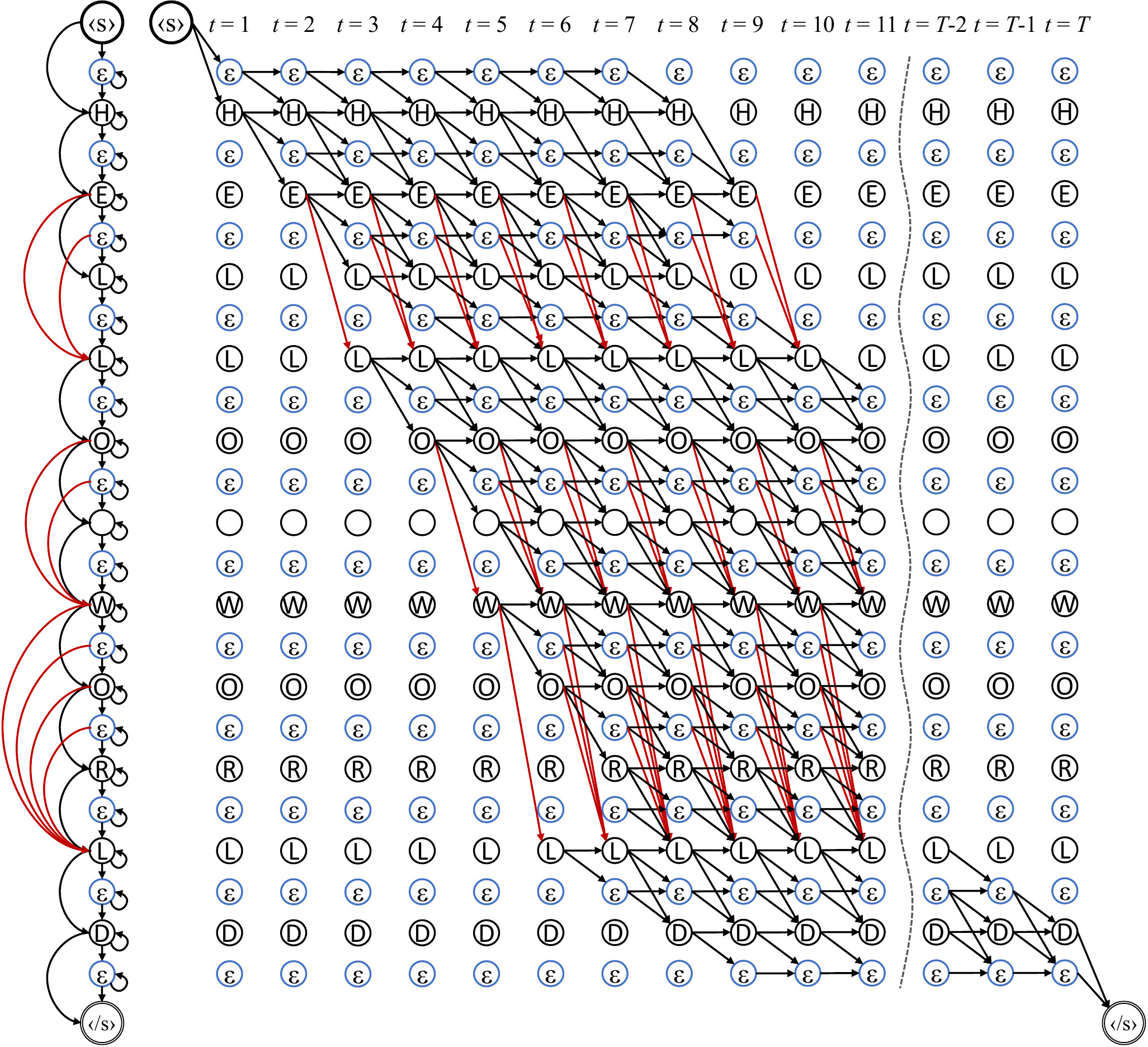}
  \caption{CTC-style confusion network $\mathcal{G}$ on the left side with the unfolded graph $\mathcal{S}(\mathcal{G},T)$ on the right side, which is generated from an $N$-best list using character-based label symbols: "HELO WORLD", "HELLO WOLD", "HELO WOLD", "HELLOWLD". Circles denote nodes with a corresponding label symbol observation inside, where $\varepsilon$ is the blank symbol. Black arrows correspond to transitions of correct label paths and red arrows to transitions of alternative label paths. Although the correct transcription is not present in the given $N$-best list, it can be recovered from the presented graph. Each transition is also associated with a transition weight, which are not shown in this figure.  }
\label{fig:graph_example}
\vspace{-0.3cm}
\end{figure}

In the supervised case, GTC can be used to train an ASR system similarly to CTC by setting the graph to insert blank symbols between the labels of the ground truth transcription and include CTC-like transition rules \cite{GravesFGS06}.
However, we here wish to consider more general cases in which there may be multiple transcriptions obtained from a seed ASR system for an unlabeled utterance.
The proposed GTC loss makes it possible to learn the model parameters with label information in a graph format. In a semi-supervised ASR scenario, we can use $N$-best hypotheses or word/token lattices as pseudo labels, which are typically given by a baseline system trained with a small amount of labeled data. As an example, a graph representation of an $N$-best hypotheses list is shown in Fig.~\ref{fig:graph_example}.
For computational reasons, it is preferable to make the graph compact while retaining correct predictions as much as possible. 

In this work, we generate a compact CTC-like graph from $N$-best hypotheses according to the following steps:
\begin{enumerate}
\item Convert the $N$-best hypotheses to a sausage-form confusion network (CN) using minimum Bayes risk decoding \cite{xu2011minimum}.%
\item Convert the CN into an optimized WFST by applying the epsilon removal, determinization, and minimization operations \cite{mohri2009weighted} to the CN, where the arc weights are operated in log-semiring to ensure the WFST remains probabilistic.
\item Convert the WFST to a CTC-like graph by replacing each state with a blank node and each arc with a non-blank node to which the arc label is assigned, where edges between the nodes are made based on the optimized WFST structure and the CTC rule, i.e., each blank node can be skipped if the adjacent non-blank nodes have different labels.
\end{enumerate}
In Step 1, a scaling factor $\mu$ can be applied to the ASR score (log probability) of each hypothesis, where the scaling factor indicates the degree to which the ASR scores are reflected in the label probabilities in the CN: $\mu=1$ means the ASR scores are used without alteration, and $\mu=0$ means the hypotheses are treated equally without considering the ASR scores.
We optionally add pruning steps after steps 1 and 2 to reduce the size of the CTC graph, which eliminate arcs if the assigned probabilities are less than a threshold $\eta$. 

\vspace{-0.1cm}
\section{Experiments}
\label{sec:experiments}
\vspace{-0.1cm}

\subsection{Dataset}
\label{ssec:dataset}
\vspace{-0.1cm}

We use as the ASR benchmark the LibriSpeech corpus of read English audio books \cite{librispeech}, which provides about 960 hours of training data, 10.7 hours of development data, and 10.5 hours of test data. The development and test data sets are both split into approximately two halves named ``clean'' and ``other'' based on the quality of the recorded utterances \cite{librispeech}. %
The training data is also split into three subsets: ``clean'' 100 hours, ``clean'' 360 hours, and ``other'' 500 hours. We use the ``clean'' 100 hours subset for supervised training and consider the remaining 860 hours as unlabeled data.

\vspace{-0.1cm}
\subsection{ASR System}
\label{ssec:wfst}
\vspace{-0.1cm}

\begin{figure}[tb]
  \centering
  \includegraphics[width=0.45\linewidth]{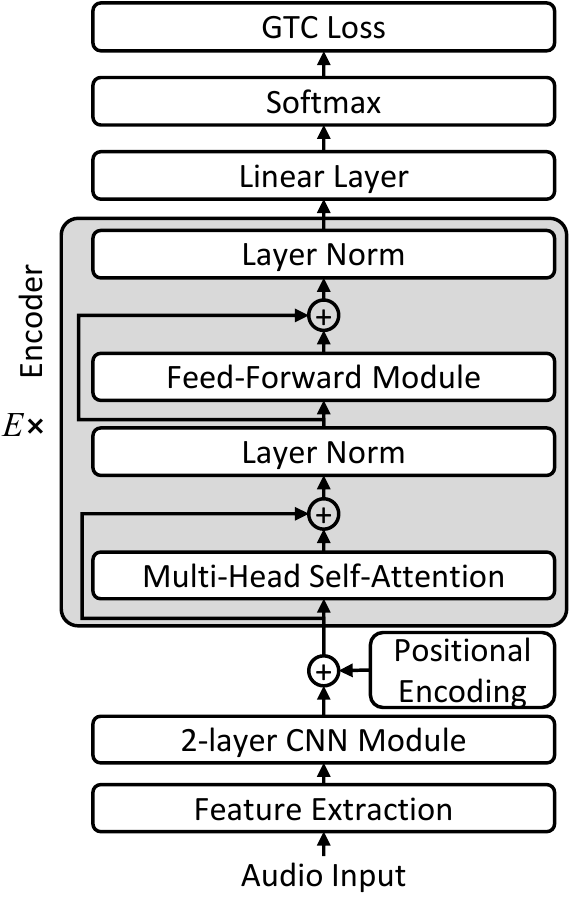}
  \caption{ ASR system architecture. }
\label{fig:nn_architecture}
\vspace{-0.3cm}
\end{figure}

Figure~\ref{fig:nn_architecture} illustrates the ASR system used in this work, a transformer-based neural network architecture that employs the proposed GTC loss function of Section~\ref{sec:gct} for training. We use 80-dimensional log-mel spectral energies plus 3 extra features for pitch information as acoustic features input to the neural network \cite{HoriWZC17}.
The neural network architecture consists of a two-layer convolutional neural network (CNN) module followed by a stack of $E=12$ transformer-based encoder layers with a linear layer plus softmax function at the end to project the neural network outputs to posterior probabilities.
Each layer of the 2-layer CNN module is using a stride of 2, a kernel size of $3 \times 3$, 320 channels, and a rectified linear unit (ReLU) activation function. In addition, a linear neural network layer is applied to the output of the last CNN layer.
Sinusoidal positional encodings \cite{VaswaniSPUJGKP17} are added to the 2-layer CNN module output before feeding it to the transformer-based encoder.
Each transformer layer employs $320$-dimensional self-attention layers with 4 attention heads, layer normalization, and a feed-forward neural network module of inner dimension $1540$. Residual connections are applied to the self-attention and feed-forward module outputs. Dropout with a probability of $0.1$ is used after self-attention and after the feed-forward module as well as for the inner dimension of the feed-forward module.
In addition, SpecAugment-based data augmentation \cite{ParkCZC19} is utilized for training. ASR output symbols consist of a blank symbol plus 5000 subwords obtained by the SentencePiece method \cite{KudoR18}, which we generated from the transcripts of the ``clean'' 100h LibriSpeech training data subset only. The ASR model is trained for 100 epochs using the Adam optimizer with $\beta_1=0.9$, $\beta_2=0.98$, $\epsilon=10^{-9}$, and learning rate scheduling similar to \cite{VaswaniSPUJGKP17} with 25000 warmup steps and a learning rate factor of $5.0$. 

A language model (LM) is employed via shallow fusion at inference time, which consists of 2 long short-term memory (LSTM) neural network layers with 1024 units each trained using stochastic gradient descent and the official LM training text data of LibriSpeech \cite{librispeech}, where we excluded sentences that occur in the 860h training data subsets.
ASR decoding is based on a time-synchronous prefix beam search algorithm similar to \cite{MaasHJN14,MoritzHR19c}. We use a decoding beam size of $30$, a score-based pruning threshold of $14.0$, an LM-weight of $0.8$, and an insertion bonus factor of $2.0$.

\vspace{-0.1cm}
\subsection{Graph Analysis}
\label{ssec:graph_analysis}
\vspace{-0.1cm}

\begin{table}[tb]
  \caption{ Oracle label error rates (LER) [\%] for the 860 hours ``unlabeled'' training data subsets ``clean'' and ``other'' using different pseudo-label representations. CN$^{20}$ denotes a confusion network generated from the 20-best ASR hypotheses for each utterance, where $_\text{low}$ and $_\text{high}$ indicate low and high pruning settings. }
  \label{tab:oracle_ER}
  \centering
     \sisetup{table-format=2.1,round-mode=places,round-precision=1,table-number-alignment = center,detect-weight=true,detect-inline-weight=math}
  \resizebox{.95\linewidth}{!}
  {\setlength{\tabcolsep}{4pt}
  \begin{tabular}{cSSSSSS}
  \toprule
  data set & \multicolumn{1}{c}{1-best} & \multicolumn{1}{c}{10-best} & \multicolumn{1}{c}{20-best} & \multicolumn{1}{c}{CN$^{20}$} & \multicolumn{1}{c}{CN$^{20}_\text{low}$} & \multicolumn{1}{c}{CN$^{20}_\text{high}$} \\
  \cmidrule(lr){1-1}\cmidrule(lr){2-7}
  clean 360h & 11.3 & 9.0 & 8.7 & 8.3 & 8.6 & 8.9 \\
  other 500h & 18.3 & 15.7 & 15.3 & 14.5 & 14.8 & 15.3 \\
\bottomrule
  \end{tabular}}
   \vspace{-1mm}
\end{table}

A seed ASR model is trained by using the 100h ``clean'' LibriSpeech training data set and an $N$-best list of ASR hypotheses is generated for each utterance of the remaining 860h of training data.
Each $N$-best list of pseudo-labels is used to generate a CTC-style confusion network (CN) as discussed in Section~\ref{sec:graph_generation} with different settings for pruning. We compare ``no'', ``low'', and ``high'' pruning settings in our experiments, where we use a scaling factor of $\mu=0.6$ and a threshold of $\eta=0.02$ for ``low'' and $\eta=0.05$ for ``high'' pruning. The pruning settings are determined based on experiments using the development data sets of LibriSpeech, where selected parameters resulted in a good trade-off between the oracle label error rates (LERs) and the graph densities, which ideally should both be small.
Pruning reduces the size and density of a graph, as can be identified by the ratio of the number of non-blank nodes in the graph to the number of labels in a reference sequence, the ground-truth transcription:
the average graph densities for ``no'', ``low'', and ``high'' pruning are 1.510, 1.233, and 1.119 for the ``clean'' 360h training data set, and 1.545, 1.275, and 1.147 for the ``other'' 500h training data set.
Table~\ref{tab:oracle_ER} shows the oracle LERs for $N$-best lists of different sizes $N$ as well as for CNs that are generated from the 20-best ASR hypotheses for each utterance.
Oracle LERs of an $N$-best list are obtained by selecting the best pseudo-label sequence from that list, i.e., the sequence that has the smallest edit distance compared to the ground-truth transcription. Oracle LERs of a graph correspond to the minimum edit distance between an FST and a reference sequence, the ground-truth transcription, which is computed by a composition operation between an acyclic FST and an edit-distance FST, followed by a single-source shortest-path algorithm \cite{mohri2003editdist}.
We can see from Table~\ref{tab:oracle_ER} that an $N$-best list contains ASR hypotheses with much lower error rates compared to the 1-best hypothesis. For example, selecting the oracle hypothesis from the 20-best list reduces the average LER by $2.5\%$ (clean 360h) and $3.0\%$ (other 500h) on an absolute scale.
Using an oracle pseudo-label computed from an $N$-best list in a graph format reduces the LER even further, since a graph representation of an $N$-best list allows for more flexible label combinations, as illustrated in Fig.~\ref{fig:graph_example}.

\vspace{-0.1cm}
\subsection{ASR Results}
\label{ssec:asr_results}
\vspace{-0.1cm}

\begin{table}[tb]
  \caption{ WERs [\%] of models trained with different pseudo-label representations. N/A denotes the seed ASR model, which is used to generate pseudo-labels. 1-best denotes self-training results using the best ASR hypothesis only.  $W=1.0$ indicates that all transition weights of the graph CN$^{20}$ are set to 1, and $W=p$ that probabilistic transition weights are used according to Section~\ref{sec:graph_generation}. }
  \label{tab:results}
  \centering
     \sisetup{table-format=2.1,round-mode=places,round-precision=1,table-number-alignment = center,detect-weight=true,detect-inline-weight=math}
  \resizebox{.95\linewidth}{!}
  {%
  \begin{tabular}{lccSSSS} %
  \toprule
   &&& \multicolumn{2}{c}{dev} & \multicolumn{2}{c}{test} \\ %
   \cmidrule(l{0.25em}r{0.25em}){4-5}\cmidrule(l{0.25em}r{0.25em}){6-7} %
  pseudo-labels&$W$&pruning & \multicolumn{1}{c}{clean} & \multicolumn{1}{c}{other} & \multicolumn{1}{c}{clean} & \multicolumn{1}{c}{other} \\ %
  \cmidrule(l{0.25em}r{0.25em}){1-3}\cmidrule(l{0.25em}r{0.25em}){4-7}
  N/A  &  &  & 7.9 & 20.0 & 8.2 & 21.2 \\ %
[0.4ex]
\hline\noalign{\vskip 0.65ex}
  1-best &&& 5.8 & 14.9 & 6.0 & 15.6 \\  %
[0.4ex]
\hdashline\noalign{\vskip 0.65ex}
  CN$^{20}$ & $1.0$ & no & 5.8 & 14.7 & 6.0 & 15.5 \\ %
  CN$^{20}$ & $p$ & low  & 5.7 & 14.7 & 5.9 & 15.1 \\ %
  CN$^{20}$ & $1.0$ & high & 5.6 & 14.4 & 6.0 & 15.2 \\  %
  CN$^{20}$ & $p$ & high & 5.6 & 14.2 & 5.9 & 15.0 \\ %
[0.4ex]
\hline\noalign{\vskip 0.65ex}
  oracle 20-best&& & 5.1 & 13.6 & 5.4 & 14.2 \\ %
  ground-truth& && 3.3 & 8.2 & 3.5 & 8.4 \\ %
\bottomrule
  \end{tabular}}
   \vspace{-1mm}
\end{table}

GTC-based ASR results for different pseudo-label representations are shown in Table~\ref{tab:results}. ``N/A'' denotes results of the seed ASR model, which is trained using the 100h of labeled clean LibriSpeech training data only.
CN$^{20}$ indicates the use of CTC-style confusion networks that are generated from the 20-best ASR hypotheses obtained from the seed model for each utterance of the 860h of unlabeled training data.
Four different CN setups are compared: 1) without pruning and with all transition weights set to 1.0, 2) with probabilistic transition weights and low pruning (see Section~\ref{ssec:graph_analysis}), 3) with transition weights set to 1.0 and high pruning, and 4) with probabilistic transition weights and high pruning. In addition, ASR results obtained when training on the best pseudo-label sequence manually selected in an oracle fashion from the 20-best list are shown as ``oracle 20-best'' and results for supervised ASR using the ground-truth transcriptions as ``ground-truth''.
Table~\ref{tab:results} shows that 1-best pseudo-labeling improves the word error rates (WERs) of the LibriSpeech test data sets by a large margin, e.g., from $21.2\%$ to $15.6\%$ for test-other. Training on multiple pseudo-label sequences encoded in graph format further improves the WERs, whereby the best results are obtained when using a graph with high pruning settings and probabilistic transitions weights.
Although an unpruned graph is more likely to contain the correct transcription, we suppose the large variance in such a graph makes it harder to learn the best label sequence due to more label noise.
Therefore, pruning and the use of transition weights can guide self-training to find pseudo-label sequences of lower error rates. 
In the best setting, the proposed GTC-based self-training approach achieves $0.7\%$ (dev-other) and $0.6\%$ (test-other) lower WERs compared to 1-best pseudo-labeling and $0.6\%$ and $0.8\%$ higher WERs compared to the ``oracle 20-best'' ASR results, which approximately correspond to a lower bound for training on multiple pseudo-labels obtained from an $N$-best list.

\vspace{-0.1cm}
\section{Conclusions}
\label{sec:conclusions}
\vspace{-0.1cm}

We proposed a new objective function that generalizes the popular CTC loss function to accept weighted finite automata in order to train with label information in a graph format with user-defined transition rules and transition weights. The proposed graph-based temporal classification (GTC) loss is applied to a semi-supervised ASR problem in order to leverage a graph with a CTC-like structure generated from an $N$-best list of pseudo-labels for self-training. We demonstrate that GTC-based self-training improves ASR results compared to 1-best pseudo-labeling. The use of pruned graphs and probabilistic transition weights further helps GTC to better exploit pseudo-label information from such a graph with improved ASR results. In our experiments, GTC-based self-training achieved up to $0.7\%$ better WERs compared to the commonly used 1-best pseudo-labeling approach, reducing the gap to an oracle experiment where the best pseudo-label sequence was selected manually from the 20-best list of ASR hypotheses.

\balance
\bibliographystyle{IEEEtran}
\bibliography{refs}

\begin{thebibliography}{10}
\providecommand{\url}[1]{#1}
\csname url@samestyle\endcsname
\providecommand{\newblock}{\relax}
\providecommand{\bibinfo}[2]{#2}
\providecommand{\BIBentrySTDinterwordspacing}{\spaceskip=0pt\relax}
\providecommand{\BIBentryALTinterwordstretchfactor}{4}
\providecommand{\BIBentryALTinterwordspacing}{\spaceskip=\fontdimen2\font plus
\BIBentryALTinterwordstretchfactor\fontdimen3\font minus
  \fontdimen4\font\relax}
\providecommand{\BIBforeignlanguage}[2]{{%
\expandafter\ifx\csname l@#1\endcsname\relax
\typeout{** WARNING: IEEEtran.bst: No hyphenation pattern has been}%
\typeout{** loaded for the language `#1'. Using the pattern for}%
\typeout{** the default language instead.}%
\else
\language=\csname l@#1\endcsname
\fi
#2}}
\providecommand{\BIBdecl}{\relax}
\BIBdecl

\bibitem{Li2020fast}
B.~Li, S.-Y. Chang, T.~Sainath, R.~Pang, Y.~R. He, T.~Strohman, and Y.~Wu,
  ``Towards fast and accurate streaming end-to-end {ASR},'' in \emph{Proc.
  ICASSP}, May 2020.

\bibitem{MoritzHLR20}
N.~Moritz, T.~Hori, and J.~Le~Roux, ``Streaming automatic speech recognition
  with the transformer model,'' in \emph{Proc. ICASSP}, May 2020, pp.
  6074--6078.

\bibitem{wang2020low}
C.~Wang, Y.~Wu, S.~Liu, J.~Li, L.~Lu, G.~Ye, and M.~Zhou, ``Low latency
  end-to-end streaming speech recognition with a scout network,'' in
  \emph{Proc. Interspeech}, Oct. 2020.

\bibitem{Pundak2016LFR_NN}
G.~Pundak and T.~Sainath, ``Lower frame rate neural network acoustic models,''
  in \emph{Proc. Interspeech}, Sep. 2016.

\bibitem{Baevski2020wav2vec}
A.~Baevski, H.~Zhou, A.~Mohamed, and M.~Auli, ``wav2vec 2.0: A framework for
  self-supervised learning of speech representations,'' \emph{arXiv preprint
  arXiv:2006.11477}, 2020.

\bibitem{Liu2020mockingjay}
A.~T. Liu, S.-w. Yang, P.-H. Chi, P.-c. Hsu, and H.-y. Lee, ``Mockingjay:
  Unsupervised speech representation learning with deep bidirectional
  transformer encoders,'' in \emph{Proc. ICASSP}, May 2020.

\bibitem{khurana2020cstnet}
S.~Khurana, A.~Laurent, and J.~Glass, ``Cstnet: Contrastive speech translation
  network for self-supervised speech representation learning,'' \emph{arXiv
  preprint arXiv:2006.02814}, 2020.

\bibitem{Hori2019cyclecon}
T.~{Hori}, R.~{Astudillo}, T.~{Hayashi}, Y.~{Zhang}, S.~{Watanabe}, and J.~{Le
  Roux}, ``Cycle-consistency training for end-to-end speech recognition,'' in
  \emph{Proc. ICASSP}, May 2019, pp. 6271--6275.

\bibitem{liu2019adversarial}
A.~H. {Liu}, H.~{Lee}, and L.~{Lee}, ``Adversarial training of end-to-end
  speech recognition using a criticizing language model,'' in \emph{Proc.
  ICASSP}, May 2019, pp. 6176--6180.

\bibitem{Lamel2002lightlySupASR}
L.~Lamel, J.-L. Gauvain, and G.~Adda, ``Lightly supervised and unsupervised
  acoustic model training,'' \emph{Comput. Speech Lang.}, vol.~16, no.~1, pp.
  115--129, 2002.

\bibitem{Novotney2009unsupAMLM}
S.~{Novotney}, R.~{Schwartz}, and J.~{Ma}, ``Unsupervised acoustic and language
  model training with small amounts of labelled data,'' in \emph{Proc. ICASSP},
  Apr. 2009, pp. 4297--4300.

\bibitem{Huang2013SemisupervisedGA}
Y.~Huang, D.~Yu, Y.~Gong, and C.~Liu, ``Semi-supervised {GMM} and {DNN}
  acoustic model training with multi-system combination and confidence
  re-calibration,'' in \emph{Proc. Interspeech}, Aug. 2013.

\bibitem{vesely2013semisup}
K.~{Veselý}, M.~{Hannemann}, and L.~{Burget}, ``Semi-supervised training of
  deep neural networks,'' in \emph{Proc. ASRU}, Dec. 2013, pp. 267--272.

\bibitem{Li2019SemiSupWeakDistil}
B.~{Li}, T.~N. {Sainath}, R.~{Pang}, and Z.~{Wu}, ``Semi-supervised training
  for end-to-end models via weak distillation,'' in \emph{Proc. ICASSP}, May
  2019, pp. 2837--2841.

\bibitem{weninger2020semisupervised}
F.~Weninger, F.~Mana, R.~Gemello, J.~Andrés-Ferrer, and P.~Zhan,
  ``Semi-supervised learning with data augmentation for end-to-end {ASR},'' in
  \emph{Proc. Interspeech}, Oct. 2020.

\bibitem{xu2020iterative}
Q.~Xu, T.~Likhomanenko, J.~Kahn, A.~Hannun, G.~Synnaeve, and R.~Collobert,
  ``Iterative pseudo-labeling for speech recognition,'' \emph{arXiv preprint
  arXiv:2005.09267}, 2020.

\bibitem{khurana2020unsupervised}
S.~Khurana, N.~Moritz, T.~Hori, and J.~L. Roux, ``Unsupervised domain
  adaptation for speech recognition via uncertainty driven self-training,''
  \emph{arXiv preprint arXiv:2011.13439}, 2020.

\bibitem{hsu2020lpriormatch}
W.-N. Hsu, A.~Lee, G.~Synnaeve, and A.~Hannun, ``Semi-supervised speech
  recognition via local prior matching,'' \emph{arXiv preprint
  arXiv:2002.10336}, 2020.

\bibitem{dey2019exploiting}
S.~Dey, P.~Motlicek, T.~Bui, and F.~Dernoncourt, ``Exploiting semi-supervised
  training through a dropout regularization in end-to-end speech recognition,''
  \emph{arXiv preprint arXiv:1908.05227}, 2019.

\bibitem{GravesFGS06}
A.~Graves, S.~Fern{\'{a}}ndez, F.~J. Gomez, and J.~Schmidhuber, ``Connectionist
  temporal classification: labelling unsegmented sequence data with recurrent
  neural networks,'' in \emph{Proc. ICML}, vol. 148, Jun. 2006, pp. 369--376.

\bibitem{liu2017gramctc}
H.~Liu, Z.~Zhu, X.~Li, and S.~Satheesh, ``Gram-{CTC}: Automatic unit selection
  and target decomposition for sequence labelling,'' in \emph{Proc. ICML}, Aug.
  2017, p. 2188–2197.

\bibitem{collobert2016wav2letter}
R.~Collobert, C.~Puhrsch, and G.~Synnaeve, ``Wav2{L}etter: an end-to-end
  {C}onv{N}et-based speech recognition system,'' \emph{arXiv preprint
  arXiv:1609.03193}, 2016.

\bibitem{manohar2018semi}
V.~Manohar, H.~Hadian, D.~Povey, and S.~Khudanpur, ``Semi-supervised training
  of acoustic models using lattice-free {MMI},'' in \emph{Proc. ICASSP}, Apr.
  2018, pp. 4844--4848.

\bibitem{Tong2019unbiasedSemiSup}
S.~Tong, A.~Vyas, P.~N. Garner, and H.~Bourlard, ``Unbiased semi-supervised
  {LF-MMI} training using dropout,'' in \emph{Proc. Interspeech}, Sep. 2019,
  pp. 1576--1580.

\bibitem{Sheikh2020}
I.~Sheikh, E.~Vincent, and I.~Illina, ``On semi-supervised {LF-MMI} training of
  acoustic models with limited data,'' in \emph{Proc. Interspeech}, Oct. 2020.

\bibitem{xu2011minimum}
H.~Xu, D.~Povey, L.~Mangu, and J.~Zhu, ``Minimum {B}ayes risk decoding and
  system combination based on a recursion for edit distance,'' \emph{Comput.
  Speech Lang.}, vol.~25, no.~4, pp. 802--828, 2011.

\bibitem{mohri2009weighted}
M.~Mohri, ``Weighted automata algorithms,'' in \emph{Handbook of Weighted
  Automata}.\hskip 1em plus 0.5em minus 0.4em\relax Springer, 2009, pp.
  213--254.

\bibitem{librispeech}
V.~{Panayotov}, G.~{Chen}, D.~{Povey}, and S.~{Khudanpur}, ``Libri{S}peech: An
  {ASR} corpus based on public domain audio books,'' in \emph{Proc. ICASSP},
  Apr. 2015.

\bibitem{HoriWZC17}
T.~Hori, S.~Watanabe, Y.~Zhang, and W.~Chan, ``Advances in joint
  {CTC}-attention based end-to-end speech recognition with a deep {CNN} encoder
  and {RNN-LM},'' in \emph{Proc. Interspeech}, Aug. 2017, pp. 949--953.

\bibitem{VaswaniSPUJGKP17}
A.~Vaswani, N.~Shazeer, N.~Parmar, J.~Uszkoreit, L.~Jones, A.~N. Gomez,
  L.~Kaiser, and I.~Polosukhin, ``Attention is all you need,'' in \emph{Proc.
  NIPS}, Dec. 2017, pp. 6000--6010.

\bibitem{ParkCZC19}
D.~S. Park, W.~Chan, Y.~Zhang, C.-C. Chiu, B.~Zoph, E.~D. Cubuk, and Q.~V. Le,
  ``{S}pec{A}ugment: {A} simple data augmentation method for automatic speech
  recognition,'' \emph{arXiv preprint arXiv:1904.08779}, 2019.

\bibitem{KudoR18}
T.~Kudo and J.~Richardson, ``Sentence{P}iece: {A} simple and language
  independent subword tokenizer and detokenizer for neural text processing,''
  \emph{arXiv preprint arXiv:1808.06226}, 2018.

\bibitem{MaasHJN14}
A.~L. Maas, A.~Y. Hannun, D.~Jurafsky, and A.~Y. Ng, ``First-pass large
  vocabulary continuous speech recognition using bi-directional recurrent
  {DNN}s,'' \emph{arXiv preprint arXiv:1408.2873}, 2014.

\bibitem{MoritzHR19c}
N.~Moritz, T.~Hori, and J.~{Le Roux}, ``Streaming end-to-end speech recognition
  with joint {CTC}-attention based models,'' in \emph{Proc. ASRU}, Dec. 2019,
  pp. 936--943.

\bibitem{mohri2003editdist}
M.~Mohri, ``Edit-distance of weighted automata: General definitions and
  algorithms,'' in \emph{Int. J. Found. Comput. Sci.}, 2003, pp. 957--982.

\end{thebibliography}

\end{document}